\documentclass{article}

\usepackage{float}
\usepackage{microtype}
\usepackage{graphicx}
\usepackage{subcaption}
\usepackage{booktabs} 

\usepackage{hyperref}

\usepackage[preprint]{icml2026}
\usepackage{amsmath}
\usepackage{amssymb}
\usepackage{mathtools}
\usepackage{amsthm}

\usepackage[capitalize,noabbrev]{cleveref}

\theoremstyle{plain}

\theoremstyle{definition}

\theoremstyle{remark}

\usepackage[textsize=tiny]{todonotes}

\usepackage{algorithm}
\usepackage{algorithmic}

\crefname{algorithm}{Alg.}{Algs.}
\Crefname{algorithm}{Algorithm}{Algorithms}

\makeatletter
\newcommand{\STATEBLOCK}[1]{%
  \STATE \parbox[t]{\linewidth}{#1}%
}
\makeatother

\usepackage{amsmath,amssymb} % for \text, \mathbb, etc.
\usepackage{mathtools}       % for \coloneqq

\DeclareMathOperator*{\argmin}{arg\,min}
\DeclareMathOperator{\BatchTopK}{BatchTopK}

\newcommand{\E}{\mathbb{E}}
\newcommand{\Laux}{\mathcal{L}_{\mathrm{aux}}}

\icmltitlerunning{Attribution-Guided Distillation of Matryoshka Sparse Autoencoders}

\begin{document}

\twocolumn[
\icmltitle{Attribution-Guided Distillation of Matryoshka Sparse Autoencoders}
  \icmlsetsymbol{equal}{*}

  \begin{icmlauthorlist}
    \icmlauthor{Cristina P. Martin-Linares}{jhu}
    \icmlauthor{Jonathan P. Ling}{jhu}
  \end{icmlauthorlist}

  \icmlaffiliation{jhu}{Department of Pathology, Johns Hopkins University, Baltimore, USA}

  \icmlcorrespondingauthor{Cristina P. Martin-Linares}{cristina.martinlinares@jhu.edu}
  %\icmlcorrespondingauthor{Jonathan P. Ling}{jling@jhu.edu}

  \icmlkeywords{Machine Learning, ICML}

  \vskip 0.3in
]

% \printAffiliationsAndNotice{\icmlEqualContribution}
\printAffiliationsAndNotice{}  

\begin{abstract}
Sparse autoencoders (SAEs) aim to disentangle model activations into monosemantic, human-interpretable features. In practice, learned features are often redundant and vary across training runs and sparsity levels, which makes interpretations difficult to transfer and reuse. We introduce \emph{Distilled Matryoshka Sparse Autoencoders} (DMSAEs), a training pipeline that distills a compact \emph{core} of consistently useful features and reuses it to train new SAEs.

DMSAEs run an iterative distillation cycle: train a Matryoshka SAE with a shared core, use gradient $\times$ activation to measure each feature’s contribution to next-token loss in the most nested reconstruction, and keep only the smallest subset that explains a fixed fraction of the attribution. Only the core encoder weight vectors are transferred across cycles; the core decoder and all non-core latents are reinitialized each time.

On Gemma-2-2B layer 12 residual stream activations, seven cycles of distillation (500M tokens, 65k width) yielded a distilled core of 197 features that were repeatedly selected. Training using this distilled core improves several SAEBench metrics and demonstrates that consistent sets of latent features can be transferred across sparsity levels.
\end{abstract}

\section*{Introduction}

Sparse autoencoders (SAEs) have become a popular method for understanding neural network activations, mapping dense model representations into sparse latent spaces that aim to encode distinct human-interpretable concepts. Despite their success, SAEs face fundamental interpretability challenges when scaled up, including feature splitting, absorption, and redundancy. For instance, a broad concept such as \emph{animal} may fragment into overly specialized features like "dog" or "cat," or a general concept may partially split, creating systematic blind spots in the representation \cite{Chanin2024}. These phenomena degrade interpretability and limit the utility of SAEs in downstream tasks \cite{Bricken2023,Gao2024}.

Matryoshka SAEs represent a recent advancement, introducing hierarchical nested dictionaries trained simultaneously to preserve general features and reduce feature absorption. This nested approach successfully retains broad concepts in smaller dictionaries and progressively specializes concepts in larger dictionaries, improving interpretability across scales \cite{Bussmann2025}. However, despite this innovation, Matryoshka SAEs still exhibit redundancy, ambiguity, and instability in their latent features, leading to interpretability and computational challenges \cite{Leask2025}. Recent work argues that mechanistic interpretability should explicitly prioritize feature consistency to support reproducibility and reuse of interpretations \cite{song2025positionmechanisticinterpretabilityprioritize,paulo2025sparseautoencoderstraineddata}.

In this work, we introduce \emph{Distilled Matryoshka Sparse Autoencoders} (DMSAEs), an attribution-guided extension of Matryoshka SAEs that distills a compact, reusable set of consistently high-value latent features. DMSAEs run a sequence of train-and-select distillation cycles. In each cycle, we train a Matryoshka SAE that reserves a shared \emph{core} of latents that are included in every prefix reconstruction. After training, we score the latents used in the most nested reconstruction (core + prefix-0) using gradient $\times$ activation attribution to the model's next-token loss, and we select the smallest subset whose cumulative attribution reaches a fixed fraction of the total attribution. We then restart training with this selected subset as the next cycle's core by copying and freezing only the encoder weight vectors, while reinitializing the decoder and all non-core latents. Iterating this cycle helps identify features that remain consistently useful across restarts, aiming to improve feature consistency across runs and sparsity levels.

\paragraph{Contributions}\mbox{}\\
(1) \emph{Attribution-guided feature distillation.} We introduce DMSAEs, an iterative train-and-select pipeline that uses gradient $\times$ activation to distill a compact set of latents (core).

(2) \emph{Feature consistency via transferable encoder directions.} By transferring and freezing only core encoder weight vectors, DMSAEs can enforce a consistent set of features across checkpoints and sparsity levels.

(3) \emph{Improved SAEBench performance.} On Gemma-2-2B layer 12 activations, we distill a 197-latent core over seven training cycles and use this core to improve on several SAEBench metrics versus Matryoshka SAE baselines.

\section*{Background}

\subsection{Sparse Autoencoders can disentangle polysemantic features}
Early mechanistic interpretability work focused on inspecting individual latents, attention heads, and MLP units. However, polysemanticity complicated these approaches because a single unit could activate for several unrelated features depending on the context \cite{Olah2020,Elhage2022}. Sparse autoencoders (SAEs) were proposed as a way to move from neuron-level to feature-level analysis by learning a sparse, overcomplete basis over model activations \cite{Bricken2023,Cunningham2023}.

Given an activation vector $x \in \mathbb{R}^d$ taken from a layer of a language model, an SAE learns an encoder and decoder
\begin{align}
f(x) &= \mathrm{ReLU}(W_{\mathrm{enc}} x + b_{\mathrm{enc}}), \label{eq:sae_encoder}\\
\hat{x} &= W_{\mathrm{dec}} f(x) + b_{\mathrm{dec}}. \label{eq:sae_decoder}
\end{align}

where $f(x) \in \mathbb{R}^m$ is constrained to be sparse and non-negative, and each dimension of $f$ is interpreted as a candidate "feature”. Variants differ primarily in how sparsity is enforced, e.g.\ $\ell_1$ penalties, TopK-style $\ell_0$ constraints, or learned thresholds \cite{Gao2024,Bussmann2024}.

This framework has enabled a wide range of interpretability applications: manual feature analysis \cite{Bricken2023}, automated feature labelling \cite{Gao2024}, circuit discovery \cite{Marks2025}, attention-head analysis \cite{Kissane2024}, and model comparison via sparse "cross-coders” \cite{Lindsey2024}. In principle, SAEs aim to map activations to sparse latents that correspond to distinct, human-interpretable features.

\subsection{Limitations of Sparse Autoencoders}
In practice, current SAEs do not always achieve this ideal. One key consideration is that SAE training minimizes a loss of the form
\[
\mathcal{L}(x) = \|x - \hat{x}\|_2^2 + \lambda \,\mathrm{sparsity}(f(x)),
\]
but only the reconstruction term is directly tied to the underlying model behavior. Sparsity regularization is a proxy for interpretability, and aggressively minimizing it produces several characteristic failure modes.

\paragraph{Feature splitting and redundancy.}
When the dictionary is large relative to the intrinsic dimensionality of the activations, sparsity pressure encourages SAEs to fragment broad concepts into many narrowly specialized latents \cite{Bricken2023}. For example, a single "punctuation marks" feature may fragment into separate latents for commas, periods, and question marks, with no latent corresponding to the more abstract category that may actually be the behaviorally relevant feature for the base model. As dictionary size increases, this fragmentation often continues, leading to many latents that are only more selective versions of existing features rather than genuinely new concepts \cite{Leask2025, Karvonen2025}.

\paragraph{Feature absorption.}
A related but more difficult to measure failure mode is feature absorption \cite{Chanin2024}. Instead of fully splitting, a general latent develops systematic gaps whenever more specific latents can take over. For instance, a latent that initially fires on numerical quantities may stop activating on dates or percentages once separate, specialized latents for dates and percentages emerge. Over time, latents affected by absorption can cover fewer cases, making them less reliable for downstream tasks and harder to interpret.

\paragraph{Feature composition and entanglement.}
Sparsity also incentivizes composite features: when two underlying factors co-occur frequently (e.g.\ color and shape), the SAE can minimize the number of active latents by representing their conjunctions ("red triangle") instead of the independent factors ("red", "triangle") \cite{Wattenberg2024}. Meta-SAE analyses suggest that many SAE latents are not truly independent. Instead, a single latent can often be approximated as a sparse mixture of a smaller set of more basic “meta-latents.” This implies that even after training an SAE, the learned features can still overlap and share structure, meaning some superposition remains in the dictionary \cite{Leask2025}. High cosine similarity among decoder vectors and the large variance explained by meta-SAEs both suggest that standard architectures have a lot of redundant and composite features \cite{Gao2024,Bussmann2024}.

\paragraph{Scaling pathologies and downstream performance.}
These problems become more severe as dictionary size grows. Larger SAEs consistently achieve better reconstruction loss \cite{Gao2024,Templeton2024}, but their performance on downstream tasks often plateaus or deteriorates \cite{Karvonen2025}. Intuitively, the model exploits the additional capacity to rearrange information in ways that are favorable for the reconstruction and sparsity objective, but worse for human-interpretable disentanglement.

Recent benchmarks reinforce a cautious view of the current state of the field. SAEBench compares many SAE architectures against simple baselines and finds that linear probes or shallow MLPs trained directly on raw activations often match or exceed the performance of SAE-derived features when labeled data is available \cite{Karvonen2025}. Thus, SAEs do not outperform other methods for supervised downstream tasks and instead introduce a unique set of trade-offs. Their clearest advantage is unsupervised feature discovery: they can reveal structure in model activations without curated labels, and resulting features can in theory be reused across analyses. \cite{Bricken2023,Templeton2024}.

In summary, architectural and objective improvements are still needed before SAEs can reliably discover stable, monosemantic features at scale.

\subsection{Matryoshka SAEs can address some SAE limitations}
Matryoshka Representation Learning (MRL) \cite{Kusupati2024} trains a single representation with a broad-to-specific structure. It optimizes losses over multiple nested prefixes of the embedding, so early coordinates capture broad features and later coordinates are more specific.

Matryoshka Sparse Autoencoders (MSAEs) adapt this idea to the SAE setting by training multiple nested dictionaries of increasing size inside a single autoencoder \cite{Bussmann2025}. MSAEs define a sequence of prefix sizes $M = \{m_1 < m_2 < \dots < m_n\}$ and optimize reconstruction losses using only the first $m_i$ latents for every $m_i \in M$. Early latents must therefore carry enough information to reconstruct the input on their own and are encouraged to capture broad, high-frequency features, while later latents can specialize in increasingly specific refinements without being allowed to absorb or overwrite the earlier ones.

When evaluated in SAEBench on Gemma-2-2B, MSAEs achieve competitive reconstruction performance while substantially reducing feature absorption, feature splitting, and feature composition relative to other SAE architectures \cite{Karvonen2025}. They also tend to perform best or near-best on sparse probing, targeted concept removal, and spurious correlation removal, and their performance improves or remains stable as dictionary size increases. These results make MSAEs a strong candidate for interpretability work that must scale to large dictionaries.

However, Matryoshka training does not fully resolve SAE pathologies. Even within an MSAE, latents at a given prefix can remain redundant, composed, or difficult to interpret. Hierarchical constraints enforce that earlier features remain general and useful, but they do not guarantee that the same high-value features are reused consistently across training runs or across sparsity levels, and they introduce additional computational cost via multiple reconstruction heads. This motivates methods that can further reduce redundancy and stabilize feature semantics on top of Matryoshka SAEs.

\section*{Distilled Matryoshka Sparse Autoencoders}
DMSAEs build on Matryoshka SAEs by reserving a small set of latent features, termed the "core", that is used in every prefix reconstruction during a training run. After each run, latents are scored by how much they contribute to the base model’s next-token loss and select a new core from the highest-scoring features. We use the word distillation to mean selecting a compact set of encoder directions that stay useful across training cycles.

\subsection*{Core placement and prefix structure.}
\begin{figure*}[!t]
  \centering
  \includegraphics[width=\textwidth]{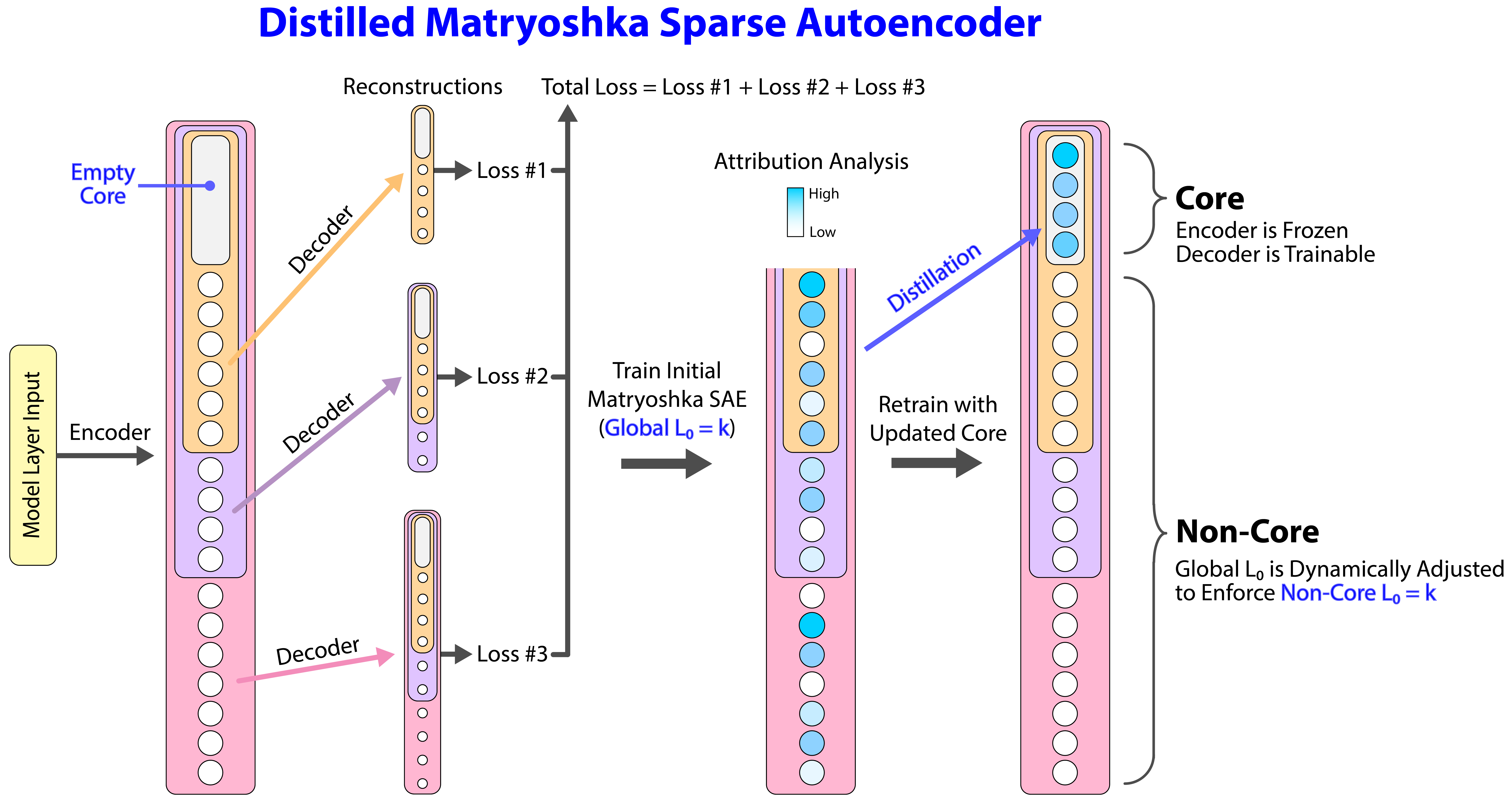}
  \caption{\textbf{DMSAE distillation schematic.} Each cycle trains a Matryoshka SAE where the smallest reconstruction uses \emph{core + prefix-0} and larger reconstructions use progressively larger non-core prefixes. After training, we score latents in \emph{core + prefix-0} using a gradient $\times$ activation score for next-token loss, and then choose the smallest set that accounts for a target fraction ($\tau$) of the total attribution. This set becomes the core for the next cycle. Across cycles, we copy and freeze only the core encoder weight vectors, while core encoder biases, core decoder weights and biases, and all non-core latents are randomly reinitialized.}
  \label{fig:dmsae_schematic}
\end{figure*}

Let the total dictionary size be $K$. We index cycles by $t=0,1,\dots,T$, where $t=0$ denotes the released SAEBench Matryoshka SAE checkpoint (not trained by us) and cycles $t\ge 1$ are the train-and-select training runs in our distillation procedure. We use the convention that $C^{(t)}$ denotes the set of latents selected after cycle $t$ by the attribution rule below (for $t=0$ this selection is computed directly from the SAEBench checkpoint) and then used as the frozen core in cycle $t{+}1$. In cycle $t{+}1$, we place the core latents as the first $|C^{(t)}|$ indices in the dictionary, followed by non-core latents partitioned into Matryoshka groups. Let $M=\{m_0<m_1<\dots<m_L\}$ denote cumulative \emph{non-core} prefix sizes. The reconstruction losses begin at \emph{core + prefix-0} (size $m_0$ of non-core), then \emph{core + prefix-1}, etc. There is no reconstruction term for the core alone. An overview of the train-and-select workflow is shown in \cref{fig:dmsae_schematic}.

\subsection*{Core and two-group sparsity}
DMSAEs split the latent dictionary into a \emph{core} and a \emph{non-core} set. Although there is no reconstruction loss for the core alone, the core contributes to every Matryoshka reconstruction during each training cycle, and its encoder weight vectors are frozen. The non-core latents remain fully trainable.

During distillation training cycles, the core is kept dense so that attribution scores are not confounded by competition for the sparsity budget. We instead apply BatchTopK only to the non-core latents so that the expected non-core $L_0$ remains approximately $k_{\text{non-core}}$ active non-core latents per token. Fixing this non-core activity budget keeps non-core representations approximately consistent across distillation cycles despite changes in core size.

\begin{equation}
\begin{aligned}
\tilde f(x;\theta,c,k_{\mathrm{NC}})
&\coloneqq
\Bigl((f(x;\theta))_{0{:}c}, \\
&\qquad \BatchTopK_{k_{\mathrm{NC}}}\bigl((f(x;\theta))_{c{:}K}\bigr)\Bigr).
\end{aligned}
\label{eq:two_group_mask}
\end{equation}

When benchmarking at a SAEBench target sparsity $k$ for a total dictionary size $K$, we scale the non-core BatchTopK target to
\begin{equation}
k_{\text{non-core}} \;=\; \operatorname{round}\!\left(k \cdot \frac{K-|C^\star|}{K}\right),
\label{eq:k_noncore}
\end{equation}

so that the fraction of non-core latents permitted to be active matches the baseline setting.
When the core is dense, the measured global activity decomposes as in Eq.~\eqref{eq:dmsae_sparsity}.

\begin{algorithm}[t]
\caption{DMSAE train-and-select distillation (dense core)}
\label{alg:dmsae_restart}
\begin{algorithmic}[1]
\REQUIRE SAEBench Matryoshka SAE checkpoint (cycle $t{=}0$); dataset $\mathcal{D}$; total width $K$;
non-core prefix sizes $M=\{m_0<\cdots<m_L\}$; cycles $T$; sparsity target $k$;
quantile $q$; attribution coverage $\tau$.
\ENSURE Selected cores $\{C^{(t)}\}_{t=0}^{T}$ and distilled core $C^\star$.
\STATEBLOCK{\textbf{Cycle 0 (no training):} set $c\leftarrow 0$,
$P^{(0)}\leftarrow\{0,\dots,m_0-1\}$, and compute
$C^{(0)}\leftarrow \textsc{SelectCoreByAttribution}(\theta^{(0)},P^{(0)},q,\tau,k,c)$.}
\FOR{$t=1$ to $T$}
  \STATE $c \leftarrow |C^{(t-1)}|$.
    \STATE \textbf{Restart: initialize $\theta^{(t)}_0$.}
    \STATE Set $W_{\mathrm{enc}}[0{:}c,:]\leftarrow W_{\mathrm{enc}}^{(t-1)}[C^{(t-1)},:]$ and freeze rows $0{:}c$.
    \STATE Reinitialize: (i) all encoder biases $b_{\mathrm{enc}}$, (ii) all non-core encoder rows $c{:}K$, (iii) the entire decoder $(W_{\mathrm{dec}},b_{\mathrm{dec}})$.
  \STATEBLOCK{$\displaystyle
\begin{aligned}[t]
&\theta^{(t)} \leftarrow {}  \argmin_{\theta\in\Theta_t}\;
\E_{x\sim\mathcal{D}}\Big[
\sum_{m\in M}\|x-\hat x_m(x;\theta,\tilde f)\|_2^2 \\
&\hspace{9.25em} {}+{} \alpha\,\Laux(x;\theta,\tilde f)
\Big],\\
&\text{where }\tilde f(x;\theta,c,k)\coloneqq {} 
\bigl((f(x;\theta))_{0{:}c},\, \\&\hspace{9.25em} \BatchTopK_{k}((f(x;\theta))_{c{:}K})\bigr).
\end{aligned}
$}
  \STATE \textbf{Reselect core:} define candidate pool $P^{(t)} = \{0,\dots,c+m_0-1\}$ (core+prefix-0).
  \STATE $\begin{aligned}[t]
    C^{(t)} \leftarrow {} & \textsc{SelectCoreByAttribution}(\theta^{(t)}, P^{(t)}, \\&\hspace{9.25em} q, \tau, k, c).
    \end{aligned}$
\ENDFOR
\STATE \textbf{Distilled core:} $C^\star \leftarrow C^{(T)} \cap C^{(T-1)}$.
\end{algorithmic}
\end{algorithm}

\Cref{alg:dmsae_restart} provides end-to-end pseudocode for the train-and-select distillation procedure, including (i) cycle indexing and core placement, (ii) within-cycle Matryoshka training with dense core, and (iii) attribution-based core replacement.
The attribution ranking and coverage-based selection rule used to form $C^{(t)}$ from the candidate pool $P^{(t)}$ are given explicitly in \Cref{alg:select_core_attr} (Appendix F).

\paragraph{Training objective (within a cycle).}
Let $\hat x_{m}(x)$ denote the reconstruction using the core latents together with the first $m$ non-core latents. Each cycle minimizes the Matryoshka reconstruction loss
\begin{equation}
\mathcal{L}_{\text{DMSAE}}(x) \;=\; \sum_{m \in M} \|x - \hat x_{m}(x)\|_2^2 \;+\; \alpha\,\mathcal{L}_{\text{aux}}(x),
\label{eq:dmsae_objective}
\end{equation}
where $\mathcal{L}_{\text{aux}}$ denotes the standard auxiliary loss used in BatchTopK-style SAEs.

\paragraph{What is transferred, what is frozen, what is reinitialized.}
At the end of cycle $t$, we obtain a selected set $C^{(t)}$ (defined by attribution below). To start cycle $t{+}1$, we initialize a new model by copying the encoder weight vectors for the selected core latents $C^{(t)}$ into the first $|C^{(t)}|$ rows of $W_{\mathrm{enc}}$ and freezing those rows during cycle $t{+}1$. We do not transfer encoder biases, and we reinitialize:
(i) all non-core encoder weights,
(ii) all encoder biases (core and non-core), and
(iii) the entire decoder (core and non-core, including $b_{\mathrm{dec}}$).
Thus the only information carried across cycles is the set of frozen encoder directions.

\paragraph{Attribution-guided core selection (what we actually compute).}
After each cycle $t$, we score each latent in the most nested reconstruction prefix (core + prefix-0) by how strongly it contributes along directions that the base model’s next-token loss is locally sensitive to.
For token position $u$, let $x_u \in \mathbb{R}^d$ be the captured activation vector (e.g.\ a residual-stream vector) and let
\[
g_u \;=\; \frac{\partial L_{\mathrm{NT}}}{\partial x_u} \in \mathbb{R}^d
\]
be the gradient of next-token cross-entropy loss with respect to that activation. 
Let $a_{u,j}$ be the SAE activation of latent $j$ at $x_u$ after applying the relevant BatchTopK masking for the candidate pool (i.e., in the smallest reconstruction prefix, core + prefix-0), and let $w^{\mathrm{dec}}_j$ be latent $j$'s decoder direction. Define its unit direction $\bar w^{\mathrm{dec}}_j = w^{\mathrm{dec}}_j/\|w^{\mathrm{dec}}_j\|_2$. We define the per-token gradient projection
\[
s_{u,j} \;=\; g_u^\top \bar w^{\mathrm{dec}}_j,
\]

and the per-token attribution magnitude (gradient $\times$ activation)
\[
\mathrm{GxA}_{u,j} \;=\; \big|a_{u,j} \, s_{u,j}\big|.
\]

To aggregate over sampled token positions, we use a high quantile (e.g.\ $q{=}0.99$) because per-token $|a_{u,j}s_{u,j}|$ is typically heavy-tailed; a high quantile emphasizes latents that achieve large attribution repeatedly while being less sensitive than a max to single outliers:

\begin{equation}
A_j \;=\; \operatorname{Quantile}_{u}\!\big(\mathrm{GxA}_{u,j}; q\big).
\label{eq:dmsae_attribution}
\end{equation}
We then sort candidate latents by $A_j$ and select the smallest set whose cumulative score reaches a target attribution coverage $\tau$:

\begin{equation}
C^{(t)} \;=\; \arg\min_{S \subseteq P^{(t)}} \left\{|S| \;:\; \sum_{j\in S} A_j \ge \tau \sum_{j\in P^{(t)}} A_j \right\}.
\label{eq:core_coverage}
\end{equation}

In practice we compute these scores over sampled token positions and then form $C^{(t)}$ by sorting and taking the smallest prefix achieving attribution coverage $\tau$; see \Cref{alg:select_core_attr} (Appendix F) for pseudocode, including the precise BatchTopK masking used for the candidate pool.

where $P^{(t)}$ is the candidate pool, defined as exactly the latents used in the smallest reconstruction loss term in cycle $t$ (the current core union prefix-0). In our experiments we use $\tau{=}0.9$.

\begin{figure*}[!t]
  \centering
  \includegraphics[width=\textwidth]{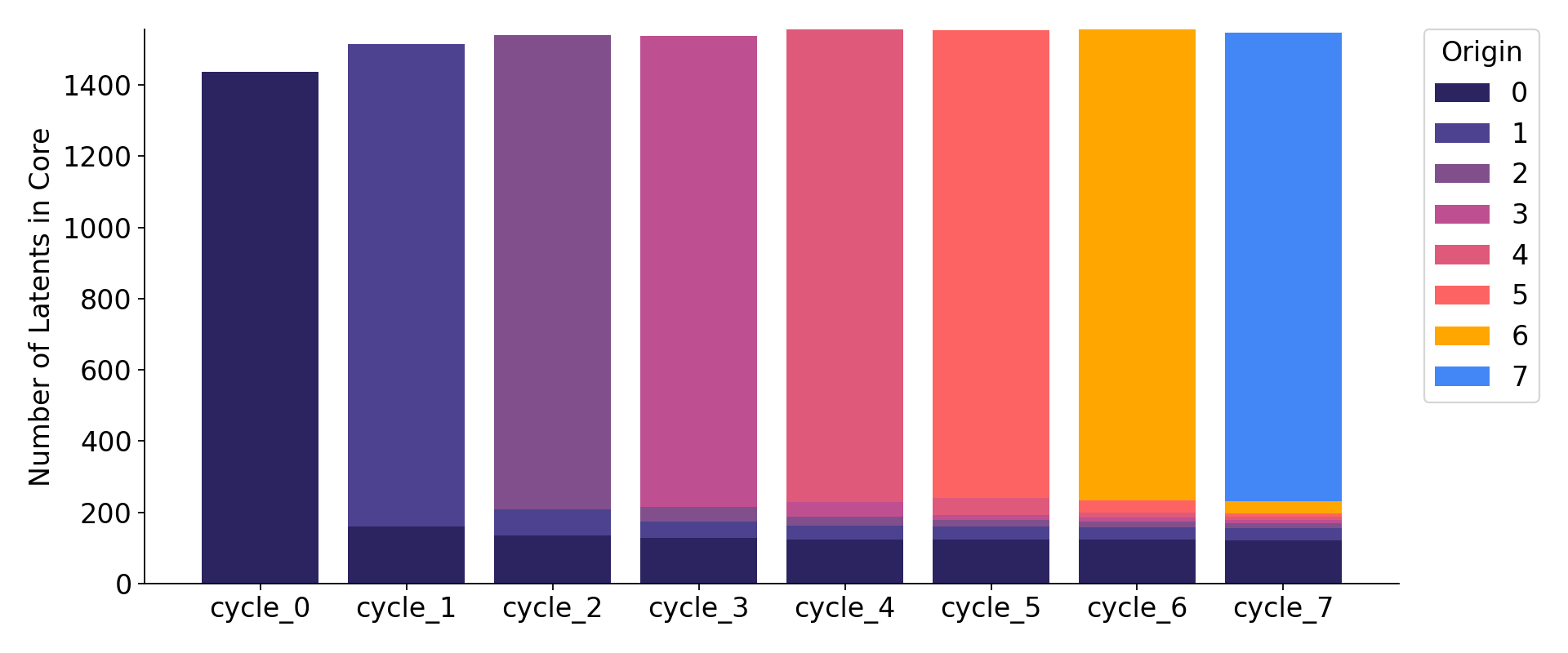}
  \caption{Evolution of core latents across distillation cycles (target sparsity $k{=}320$, selection threshold $\tau{=}0.9$). After each training cycle $t$ (including $t=0$, the SAEBench checkpoint), we rank latent features by their contribution to the language model’s next-token loss and select the smallest set $C^{(t)}$ whose cumulative score reaches 90\% of the total attribution. The y-axis shows the number of latents selected into the core at the end of cycle $t$. Colors indicate when a feature was first selected (the earliest cycle in which it appears in any $C^{(t)}$), showing how later cycles combine previously discovered latents with newly identified ones. For benchmarking, we defined the distilled core as the intersection of the final two cycle cores, yielding 197 latents.}
  \label{fig:core_latent_origins_0-1-2-3-4-5-6-7}
\end{figure*}

\paragraph{Two-stage training procedure: core distillation and distilled-core transfer}\mbox{}\\
The full train-and-select core distillation stage corresponds to running \Cref{alg:dmsae_restart} for $T$ cycles at fixed non-core sparsity (e.g.\ $k{=}320$), while the distilled-core transfer stage (training new DMSAEs at multiple target sparsities using the fixed $C^\star$) is summarized in \Cref{alg:dmsae_core_transfer} (Appendix F).

\emph{Core distillation.}
We run $T$ train-and-select cycles at a fixed non-core sparsity (here $k{=}320$). In each cycle, we train a DMSAE, compute attribution scores for latents in the smallest reconstruction prefix (core + prefix-0), and define the next cycle’s core $C^{(t)}$ as the smallest subset whose cumulative attribution reaches the target coverage. Repeating this yields a sequence of candidate cores $\{C^{(t)}\}$ (Fig.~\ref{fig:core_latent_origins_0-1-2-3-4-5-6-7}).

\emph{Distilled-core selection and transfer.}
We then extract a compact set of latents from cycle 7 that persisted across training cycles. In our experiments, we define the distilled core as the subset of latents from cycle 7 that persisted through successive training cycles,

\begin{equation}
C^\star \coloneqq C^{(T)} \cap C^{(T-1)}.
\label{eq:distilled_core}
\end{equation}

yielding 197 latents. We train new DMSAEs from scratch at multiple sparsity targets, freezing only the encoder weight vectors indexed by $C^\star$ and reinitializing all remaining parameters; these models are the ones evaluated by SAEBench.

\paragraph{Core/non-core sparsity variants.}
Our main results use the \emph{dense core} DMSAE variant; Appendix E reports a
\emph{sparse core} ablation for direct sparsity-matched comparison to SAEBench.

\emph{Dense core variant.}
We apply BatchTopK only to non-core latents, using the two-group masking rule in
Eq.~\eqref{eq:two_group_mask}. For SAEBench plots indexed by a target sparsity $k$
and total dictionary size $K$, we use the scaled non-core target $k_{\text{non-core}}$
defined in Eq.~\eqref{eq:k_noncore}.

The effective expected global sparsity is
\begin{equation}
L_{0,\mathrm{global}} \;=\; L_{0,\text{core}} \;+\; L_{0,\text{non-core}}.
\label{eq:dmsae_sparsity}
\end{equation}
In the dense core variant, BatchTopK calibration keeps
$L_{0,\text{non-core}} \approx k_{\text{non-core}}$ by construction, and
$L_{0,\text{core}}$ is measured empirically.

\emph{Sparse core variant.}
We apply the standard global BatchTopK constraint (target $k$) across the full dictionary
(core included). The only architectural difference from vanilla MSAE is that the core
encoder weight vectors are frozen.

\begin{figure*}[!t]
  \centering
  \includegraphics[width=\textwidth]{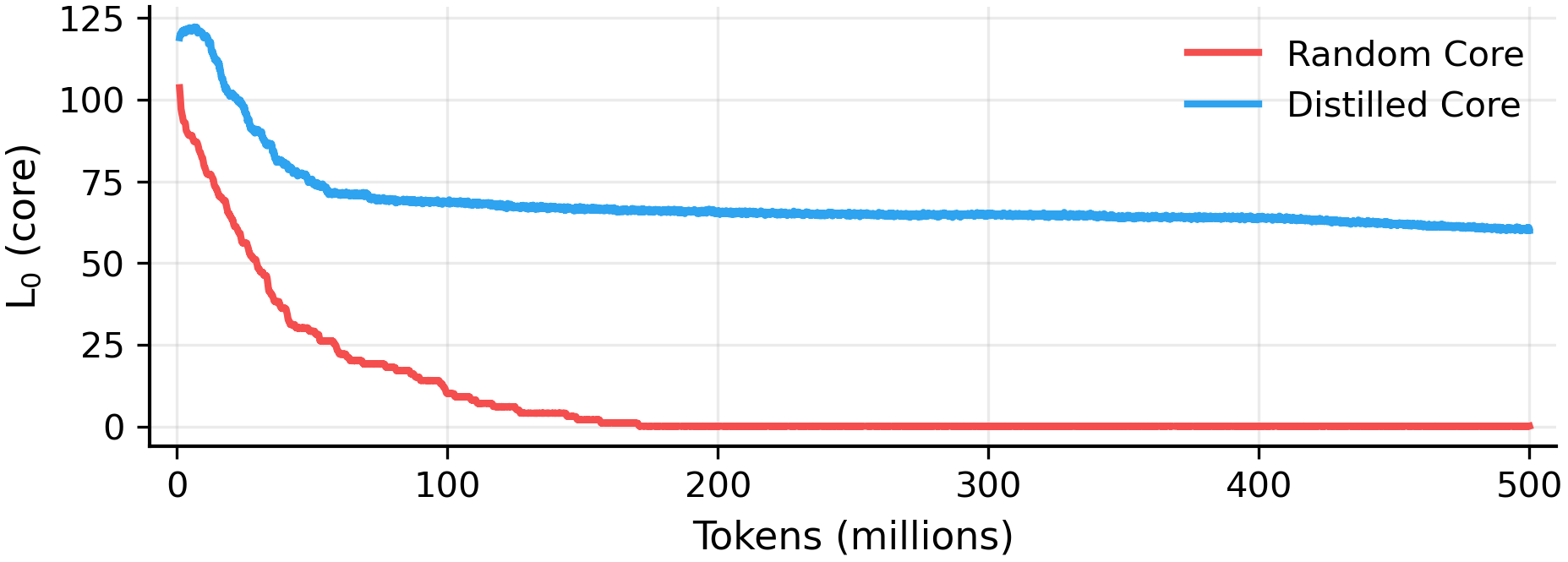}
  \caption{Distillation uncovers useful latents. We trained two DMSAEs ($k=320$), one that uses a core derived from our distillation procedure (blue; Fig.~\ref{fig:core_latent_origins_0-1-2-3-4-5-6-7}, cycle 7) and another that uses a randomly initialized core of equivalent size (red). The y-axis shows the mean number of core latents active per sample during training ($L_{0,\mathrm{core}}$). The model with a randomly initialized core drives $L_{0,\mathrm{core}}$ towards zero during training, while the model with the distilled core continues to rely on the core throughout training. This suggests that distillation identifies core latents that are systematically useful for reconstruction.}
  \label{fig:distilled_vs_random_core}
\end{figure*}

\section*{Experiments}
We evaluate Distilled Matryoshka SAEs (DMSAEs) on the SAEBench interpretability benchmark suite using activations from layer 12 of Gemma-2-2B. Our experiments proceed in three stages: (1) selecting a sparsity level and promotion threshold that yield stable cores under distillation, (2) running a longer distillation schedule at the chosen hyperparameters to obtain a compact "distilled core”, and (3) benchmarking DMSAEs built from this distilled core against the Matryoshka SAEs in SAEBench. Throughout, we follow the SAEBench evaluation protocol and report the same set of metrics.

\paragraph{Selecting sparsity and promotion threshold.}
\label{sec:experiments_promotion}

We sweep over target sparsities $k \in \{20,40,80,120,\dots,640\}$ and run four train-and-select \emph{training} cycles per setting (cycles $1$-$4$), starting from the SAEBench Matryoshka SAE checkpoint (cycle $0$). In each cycle, we compute gradient $\times$ activation attribution scores as defined in Eq.~\eqref{eq:dmsae_attribution} and form an attribution coverage curve: the cumulative fraction of cumulative attribution score captured by the top-ranked latents.

We first sweep the attribution coverage threshold ($\tau$) from $\tau{=}0.1$ to $\tau{=}1.0$ and find that promoting more latents steadily reduces reconstruction loss and increases fraction of variance explained, with $\tau{=}0.9$ close to $\tau{=}1.0$ (\cref{fig:promotion_fraction_sweep}).

Fixing the coverage threshold at $\tau{=}0.9$, we then compare cross-cycle carryover of selected core latents as a function of sparsity. As shown in \cref{fig:stacked_core_latent_origins}, very low sparsities ($k=20$-$80$) yield little carryover, while very large $k$ (a weak sparsity constraint / high activity, e.g.\ $k{=}640$) can lead to degenerate optimization behavior (e.g.\ collapse at $k{=}640$). Intermediate sparsities exhibit substantially higher reuse across cycles; we therefore focus subsequent experiments on this intermediate regime and use $k{=}320$ in all distillation runs reported below.

\paragraph{Multi-cycle distillation and the distilled core.}

Having identified $k{=}320$ and an attribution coverage threshold $\tau{=}0.9$ as our working hyperparameters, we run $T{=}7$ train-and-select \emph{training} cycles (cycles $1$-$7$) on 500M tokens from The Pile. We denote by \texttt{cycle\_0} the released SAEBench Matryoshka SAE checkpoint (not trained by us) and compute an initial promoted core $C^{(0)}$ from this checkpoint using Eq.~\eqref{eq:dmsae_attribution} to seed \texttt{cycle\_1}. Each subsequent cycle is a restart: we transfer and freeze only the selected \emph{encoder weight vectors} for the next cycle’s core, and we reinitialize all non-core parameters and the entire decoder.

At the end of each cycle $t$, we score latents in the smallest reconstruction prefix (core + prefix-0) using the gradient $\times$ activation attribution in Eq.~\eqref{eq:dmsae_attribution}, sort latents by attribution, and select the smallest set $C^{(t)}$ whose cumulative attribution reaches 90\% of the total attribution score. \cref{fig:core_latent_origins_0-1-2-3-4-5-6-7} summarizes how the selected sets evolve across cycles and shows that, while many latents are newly selected in each cycle, a subset persists across multiple restarts.

After completing $T{=}7$ cycles, we extract a compact \emph{distilled core} by retaining only latents that persist through successive training cycles. We define the distilled core $C^\star$ as in Eq.~\eqref{eq:distilled_core}.

However, to first confirm that $C^\star$ reflects genuinely useful encoder directions, we repeat the same procedure with a randomly initialized core of the same size. While the non-core BatchTopK budget is fixed by construction (Eq.~\eqref{eq:k_noncore}), the random-core model drives $L_{0,\mathrm{core}}$ toward zero during training, whereas the distilled-core model continues to rely on the core for reconstruction (\cref{fig:distilled_vs_random_core}). This indicates that attribution-guided distillation is essential for obtaining a core that remains useful under retraining.

\begin{figure*}[!t]
  \centering
  \includegraphics[width=\textwidth]{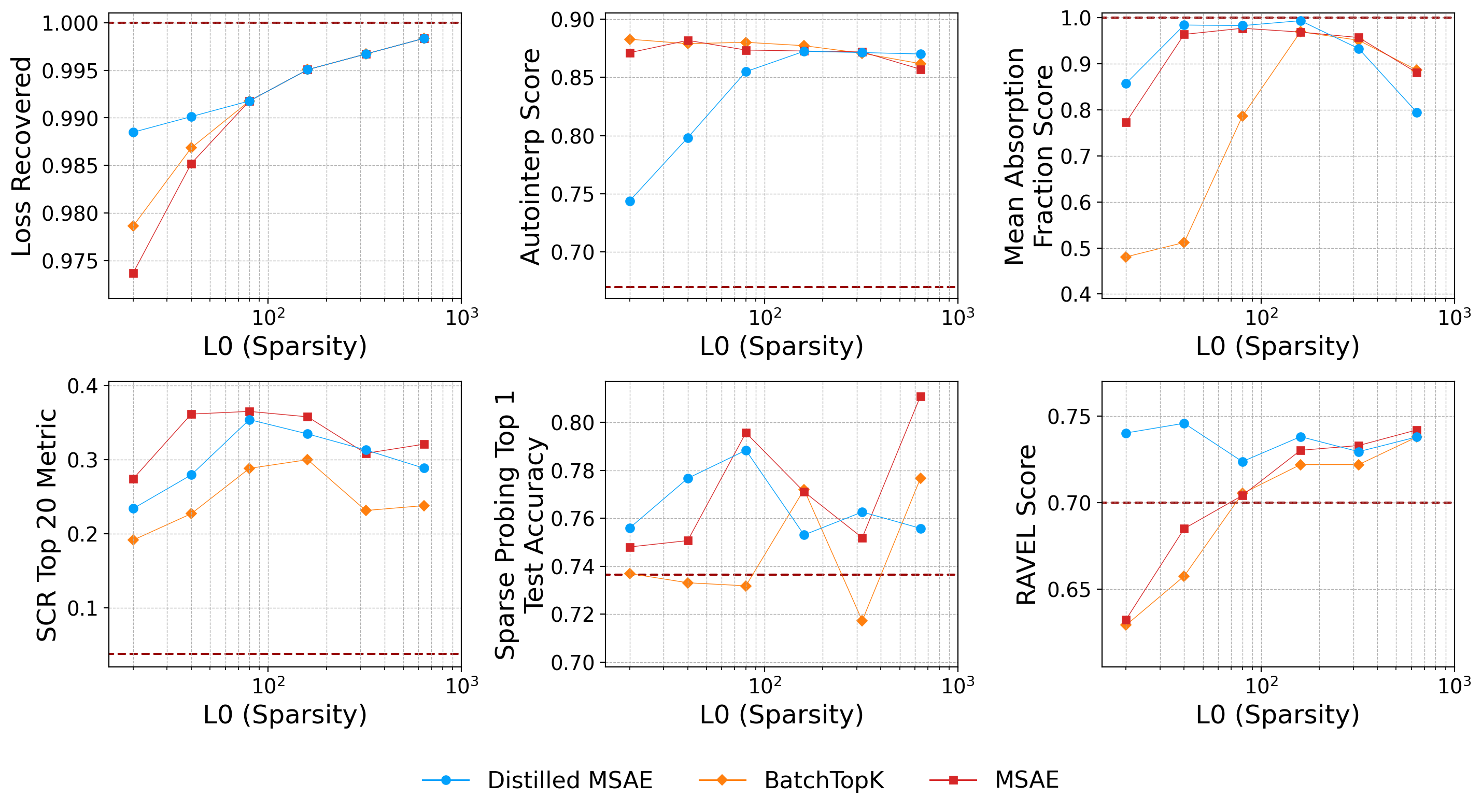}
  \caption{SAEBench performance across sparsities (dense core DMSAE). SAEBench metrics versus measured sparsity (mean number of active latents) for Distilled MSAE (DMSAE) with a dense core, compared with Matryoshka SAE (MSAE) and BatchTopK, evaluated on Gemma-2-2B layer 12 activations (all SAEs have a dictionary size of 65k). Here, the DMSAE core (197 latents) is taken from the distilled core produced by the $k{=}320$ distillation run (\cref{fig:core_latent_origins_0-1-2-3-4-5-6-7}).
  We then used this core to train new DMSAEs from scratch at each target sparsity ($k\in\{20,40,80,160,320,640\}$). Core encoder weights are frozen while sparsity is enforced on the remaining non-core latents. DMSAE matches or improves several metrics, particularly absorption and RAVEL, but AutoInterp drops sharply at the lowest sparsities. For a more direct comparison to the SAEBench baselines, Appendix E reports a sparsity-matched setting in which BatchTopK is applied globally (including the core); these runs exhibit similar overall trends (Fig.~\ref{fig:saebench_core_sparsity_ablation}).}
  \label{fig:saebench_main}
\end{figure*}

\paragraph{Building DMSAEs from the distilled core.}
Using the distilled core $C^\star$ (197 latents), we train new models with random initialization at each target sparsity: we initialize all non-core parameters randomly, initialize the core encoder weight vectors from the distilled core, and freeze only these core encoder weights. Thus, the training cost for each model matches the SAEBench baselines and the computationally expensive multi-cycle training procedure is used only to identify $C^\star$ at $k=320$.

For SAEBench benchmarking we evaluate two sparsity regimes. In the \emph{dense core} regime (used in the main paper), we apply top $k$ only to the non-core latents with the proportionally scaled integer with $k_{\text{non-core}}$ set by Eq.~\eqref{eq:k_noncore}, and do not sparsify the core. In the \emph{sparse core} regime (Appendix E), we apply standard global top $k$ across the entire dictionary (core included), which is otherwise identical to the vanilla MSAE setup except for the frozen core encoder weight vectors.

\Cref{fig:saebench_main} summarizes SAEBench metrics for DMSAEs and Matryoshka SAEs across sparsities.

We summarize this fixed-core training procedure (including the dense core $k_{\text{non-core}}$ scaling in Eq.~\eqref{eq:k_noncore} and the sparse core ablation) in \Cref{alg:dmsae_core_transfer} (Appendix F).

\paragraph{Sparsity matching.}
Our primary SAEBench results use the dense core DMSAE variant, where sparsity is only enforced on the non-core latents. We use this setting for core discovery because allowing the core to activate freely maximizes its opportunity to contribute to reconstruction and yields a more reliable attribution signal for promotion. A second motivation is conceptual: the core encoder directions are distilled from models trained under sparse coding, so the transferred encoder vectors already represent features that are monosemantic and non-redundant. Thus, we treat the repeatedly selected core as a small set of validated encoder directions and allow them to fire whenever relevant, rather than force them to compete with newly learned non-core features for the sparsity budget.

Dense core models are therefore not strictly sparsity-matched to vanilla Matryoshka SAEs at the same nominal target $k$, because the measured global $L_0$ also includes core activity. To make a more explicit comparison, we have additionally trained sparse core DMSAEs that apply the standard global sparsity constraint across the full dictionary (core included). These sparse core DMSAEs differ from vanilla MSAE only in the freezing of the core encoder weight vectors (Appendix E; \cref{fig:saebench_core_sparsity_ablation}) and follow similar overall trends compared to the core dense DMSAEs.

\section*{Limitations}
The DMSAE method requires multiple cycles of distillation to identify a stable core, leading to much higher computational costs relative to training a single Matryoshka SAE. However, once a high-value core has been identified, training and evaluating DMSAEs with a fixed core costs the same as training a standard Matryoshka SAE. It is likely that useful latents can be identified more efficiently than via multi-cycle distillation. 

Core selection depends on the attribution method. Gradient $\times$ activation is a fast first-order approximation, but may be suboptimal. Latent ablation offers a more direct measure of importance, but at higher computational cost.

Freezing encoder weight vectors stabilize each core latent’s detection direction, which largely determines what it responds to. However, because the core encoder biases and decoder weights and biases are reinitialized, a latent’s threshold, firing rate, and contribution to reconstruction can still change across cycles. Thus, DMSAEs enforce consistency at the level of detection directions, but do not guarantee identical latent behavior across retrainings.

Finally, we do not enforce sparsity on the distilled core during transfer, because these directions were selected under sparse training and are intended to remain available whenever they are useful. This can increase the total number of active latents, so comparisons can depend on how sparsity is defined. Therefore, \cref{fig:saebench_core_sparsity_ablation} reports a sparsity-matched control (global BatchTopK that includes the core) for direct comparison to baseline Matryoshka SAEs, and these results show the same qualitative trends as dense core DMSAEs.

\section*{Conclusion}
We introduce Distilled Matryoshka Sparse Autoencoders (DMSAEs), which use an attribution-guided training strategy to distill a compact core of reusable latent features from Matryoshka SAEs. By transferring and freezing only core encoder directions across cycles, DMSAEs promote a consistent set of high-value features. This provides a practical route toward feature consistency across SAEs and transferability of interpretations across training runs and sparsity levels. On Gemma-2-2B layer 12 activations, distillation yields a 197-latent core and improves multiple SAEBench metrics compared to Matryoshka SAE baselines. Future work focuses on reducing the computational cost of distillation and extending the DMSAE framework to other models and scientific domains.

\bibliography{references}
\bibliographystyle{icml2026}

\newpage
\appendix
\onecolumn

\section*{A. Core and non-core sparsity: formal statements}
This appendix records simple bookkeeping facts for the two-group sparsity scheme used by DMSAEs.

Let $C$ and $N$ denote the index sets of core and non-core latents. For an input $x$, let $f(x)\in\mathbb{R}^K$ be the post-nonlinearity latent vector (before any top-$k$ / BatchTopK masking). Let $\tilde f(x)$ denote the latent vector \emph{actually used for reconstruction} after applying the relevant sparsity mechanism: in dense core variants we apply BatchTopK only to the non-core coordinates, while in sparse core variants we apply BatchTopK globally across all $K$ latents. We define the active sets using $\tilde f$:
\[
S_C(x) = \{ j\in C : \tilde f_j(x) > 0 \},
\qquad
S_N(x) = \{ j\in N : \tilde f_j(x) > 0 \}.
\]

We denote the corresponding average sparsities by
\[
L_{0,\text{core}} = \mathbb{E}_x |S_C(x)|,
\qquad
L_{0,\text{non-core}} = \mathbb{E}_x |S_N(x)|.
\]

\paragraph{Remark (Non-core budget calibration).}
In the dense core variant, BatchTopK is applied only to the non-core coordinates and is calibrated to a fixed target $\bar L_{0,\text{non-core}}$ (equivalently, a fixed $k_{\text{non-core}}$). Consequently, the expected non-core activity $L_{0,\text{non-core}}$ is approximately invariant across distillation cycles even as the core size changes; the global activity $L_{0,\mathrm{global}}$ can still vary due to changes in $L_{0,\mathrm{core}}$.

 \paragraph{Proposition 2 (Global $L_0$ decomposition).}
Under the definitions above, the expected global sparsity decomposes as in Eq.~\eqref{eq:dmsae_sparsity}.

\emph{Proof.} $S_C(x)$ and $S_N(x)$ are disjoint by construction, so
$|S_C(x)\cup S_N(x)| = |S_C(x)|+|S_N(x)|$. Taking expectations yields Eq.~\eqref{eq:dmsae_sparsity}.
\hfill$\square$

These identities justify Eq.~\eqref{eq:dmsae_sparsity}: in dense core variants, the effective global sparsity increases by the observed $L_{0,\mathrm{core}}$ while keeping the non-core budget fixed.

\section*{B. Connections to structured sparsity and multi-task feature learning}
This appendix situates DMSAEs relative to several classical viewpoints in representation learning and structured regularization. These connections are intended as conceptual guides rather than formal equivalences, since DMSAEs use a nonconvex training objective and a BatchTopK-style activation constraint.

\paragraph{Shared-plus-sparse decomposition (multi-task / "dirty model").}
The DMSAE core can be viewed as a shared representation reused across the family of reconstruction problems induced by different Matryoshka prefixes and distillation cycles. Each prefix length defines a related task, and DMSAEs reuse a common subset of latents while allowing the remaining latents to specialize. This resembles multi-task feature learning and "dirty model" formulations in which parameters for multiple related tasks are decomposed into a shared component plus a sparse task-specific component \cite{Jalali2010,YangRavikumar2013}. Under appropriate incoherence and sparsity assumptions, these models admit guarantees for support recovery and bounded estimation error, motivating the intuition that a small shared set of features can be beneficial when tasks share structure.

\paragraph{Hierarchical and structured sparsity.}
Requiring that core latents participate in every Matryoshka prefix induces a simple hierarchical structure: the core provides a mandatory base representation, and additional (non-core) latents provide refinements as the prefix grows. This is related in spirit to hierarchical and tree-structured sparsity constraints, where activation patterns are restricted so that "child" variables can be active only when their "ancestor" variables are active \cite{Zhao2009,Jenatton2011}. In such settings, structured penalties encourage coarse-to-fine representations, with high-level groups capturing broad structure and lower-level groups capturing specialized refinements. DMSAEs induce an analogous coarse-to-fine decomposition via architecture (shared core + growing non-core prefix) rather than via an explicit tree-structured penalty.

\paragraph{Backbone-plus-residual perspective (low-rank plus sparse).}
A complementary analogy comes from low-rank plus sparse decompositions such as robust PCA, where a data matrix is decomposed into a low-rank component capturing global structure and a sparse component capturing localized deviations \cite{Candes2011}. Conceptually, the DMSAE core plays the role of a dense "backbone" of globally useful, high-frequency directions, while the non-core latents act as a sparser residual mechanism that corrects reconstruction errors and captures more specialized structure. This perspective helps explain why allowing the core to activate more frequently can be beneficial without sacrificing the effective capacity allocated to the non-core dictionary.

\paragraph{Group-specific $\ell_0$ constraints and the dynamic global $L_0$ schedule.}
The dynamic global $L_0$ rule in Eq.~\eqref{eq:dmsae_sparsity} can be interpreted as a practical approximation to a group-specific sparsity constraint that treats core and non-core latents differently. Ideally, one would like to solve an optimization problem of the form
\[
\min_\theta \ \mathbb{E}_x\big[\mathcal{L}_{\mathrm{recon}}(x;\theta)\big]
\quad \text{subject to} \quad
\mathbb{E}_x\big[\|f_N(x;\theta)\|_0\big] \leq \bar L_{0,\text{non-core}},
\]
where $f_N(x)$ denotes the non-core activations and no analogous constraint is imposed on the core activations $f_C(x)$. This corresponds to a hard budget on the non-core group while leaving the core effectively unpenalized. Related convex formulations appear in sparse-group and dirty-model regularization, where shared and task-specific components are penalized with different norms or strengths \cite{Jalali2010,Obozinski2010,YangRavikumar2013}.

In practice, our main dense core DMSAE variant applies BatchTopK only to the non-core latents, leaving the distilled core free to activate. Operationally, this keeps the non-core capacity approximately fixed across cycles while allowing the core to contribute whenever relevant, so the effective global activity is $L_{0,\mathrm{global}} = L_{0,\text{core}} + L_{0,\text{non-core}}$ (Eq.~\eqref{eq:dmsae_sparsity}). For direct comparability to SAEBench, Appendix E additionally reports a sparse core ablation that applies BatchTopK globally across all latents (core included).

\clearpage
\section*{C. Choosing the attribution coverage threshold for core promotion}
\begin{figure}[H]
  \centering
  \includegraphics[width=\textwidth,height=0.78\textheight,keepaspectratio]{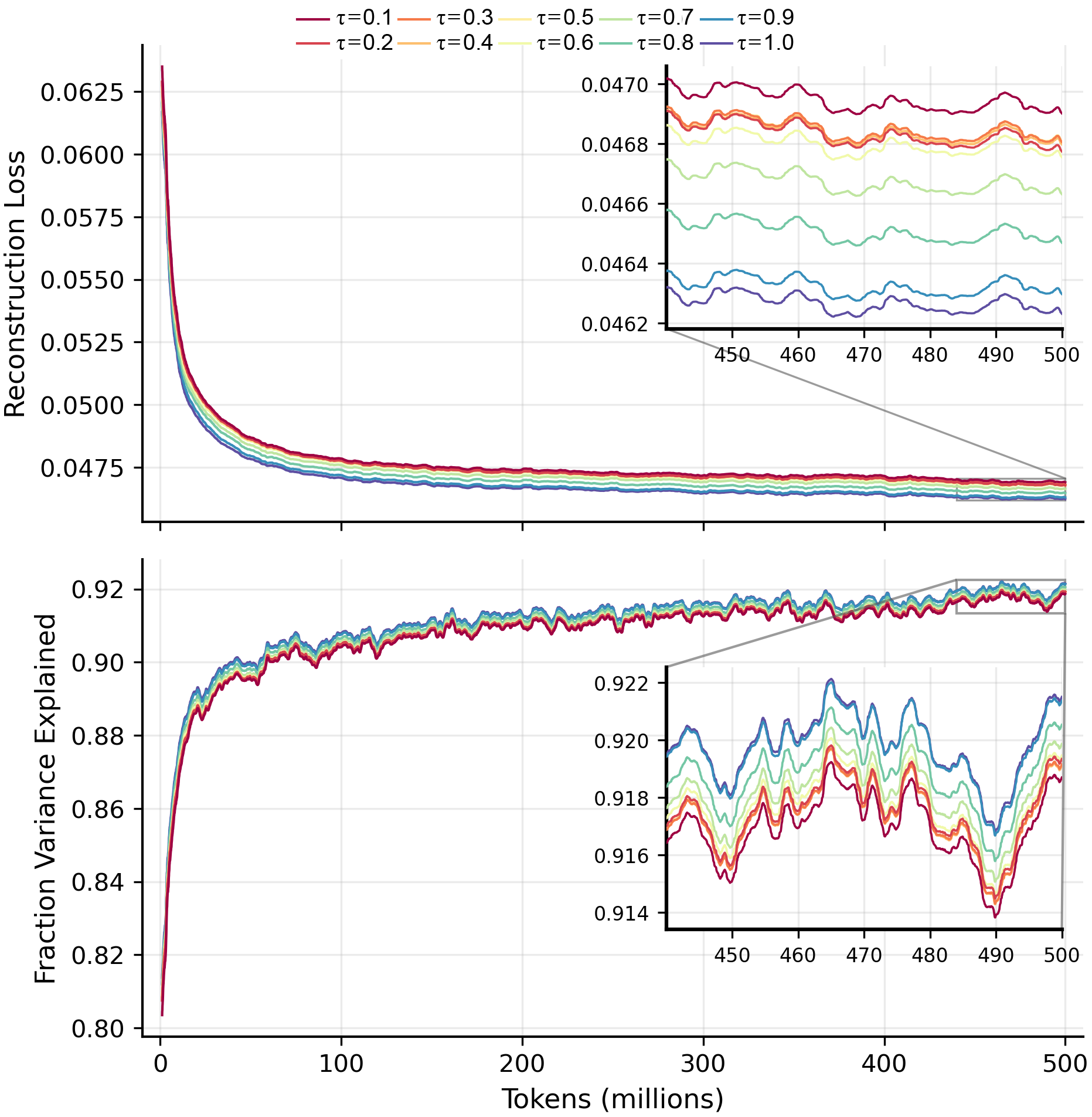}
  \caption{Choosing the attribution coverage threshold for core promotion. We trained DMSAEs ($k{=}320$) using different attribution coverage thresholds ($\tau$). Promoting more latents steadily reduces reconstruction loss (top) and increases fraction of variance explained (bottom) throughout training, with no clear saturation up to $\tau{=}1.0$. The $\tau{=}0.9$ setting nearly overlaps $\tau{=}1.0$ in fraction of variance explained and achieves similar reconstruction loss. These results justify using $\tau{=}0.9$ as our default attribution threshold.}
  \label{fig:promotion_fraction_sweep}
\end{figure}

\clearpage
\section*{D. Core stability depends strongly on the sparsity target.}
\begin{figure}[H]
  \centering
  \includegraphics[width=\textwidth,height=0.78\textheight,keepaspectratio]{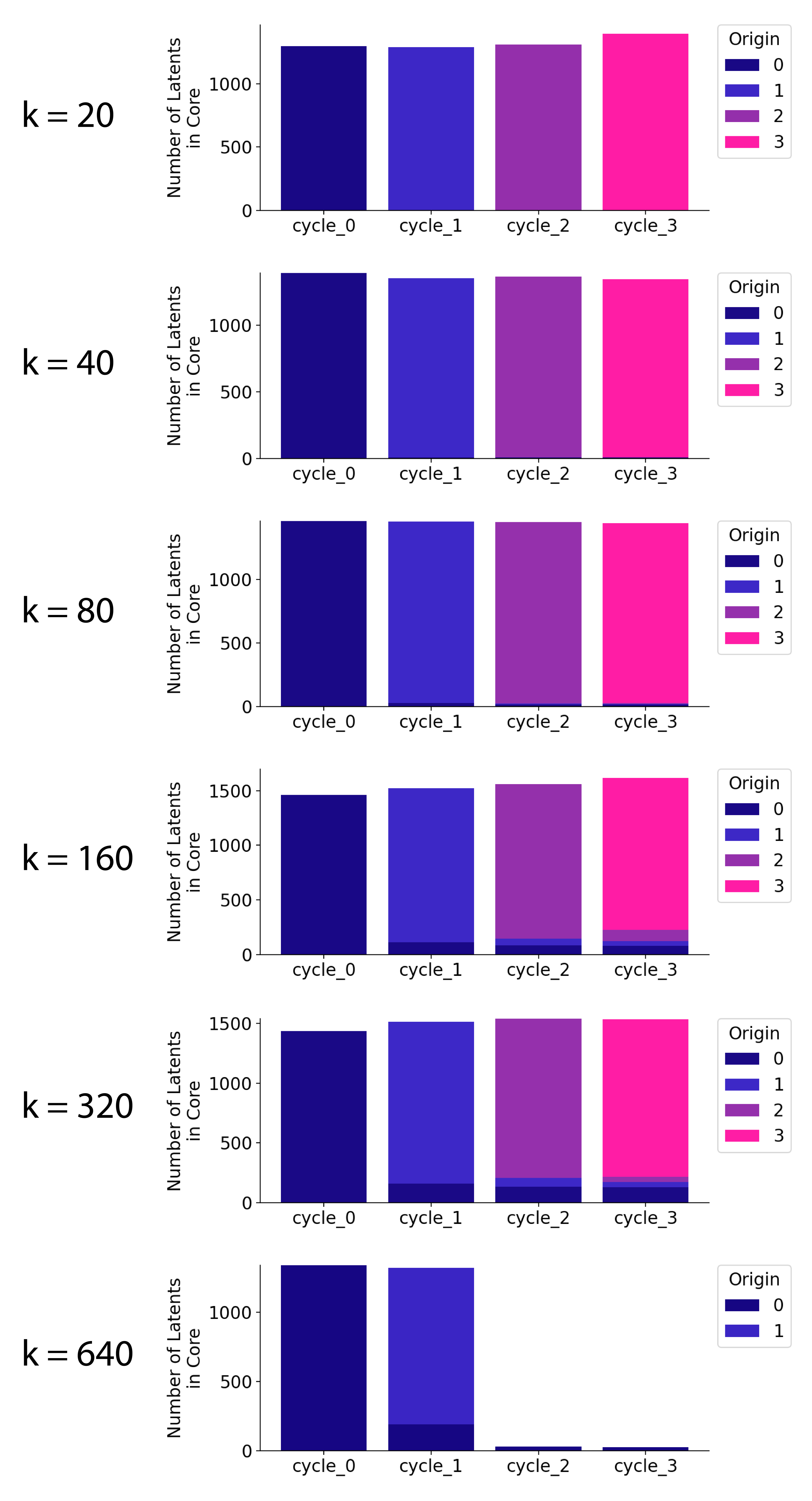}
  \caption{We ran four DMSAE distillation cycles for each top $k$ setting. After each cycle $t$, we score latents in the smallest reconstruction prefix (core + prefix-0) by gradient $\times$ activation attribution to next-token loss and select the smallest set whose cumulative score reaches 90\% of the total attribution score ($\tau{=}0.9$) as the next cycle’s core. Bars show the number of selected core latents per cycle and colors indicate the earliest cycle in which each latent was first selected (origin), visualizing carryover across cycles. Only the intermediate values ($k{=}160$ and $k{=}320$) exhibit substantial carryover into later cores. At $k{=}640$ (largest $k$, i.e., weakest sparsity constraint / highest activity), optimization degenerated after cycle 1 and reconstruction failed, thus we omitted subsequent cycles for $k{=}640$.}
  \label{fig:stacked_core_latent_origins}
\end{figure}

\clearpage
\section*{E. Core sparsity ablation on SAEBench.}
\begin{figure}[H]
  \centering
  \includegraphics[width=\textwidth,height=0.78\textheight,keepaspectratio]{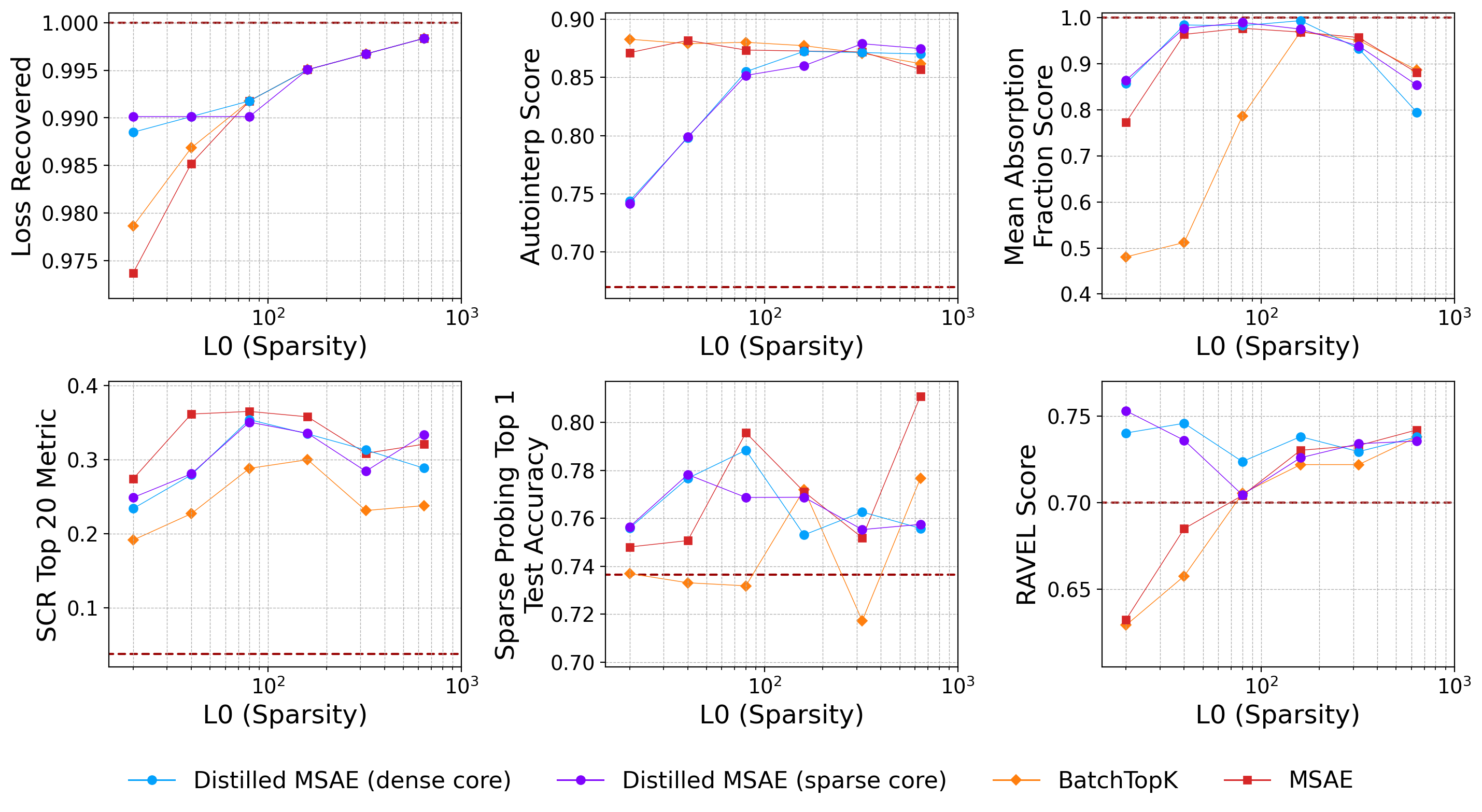}
  \caption{Performance across top k targets in the SAEBench (as in Fig.~\ref{fig:saebench_main}), but comparing two DMSAE sparsity rules using the same distilled core (197 latents). Dense core (blue) applies BatchTopK sparsity only to the non-core latents and leaves the distilled core free to activate. Sparse core (purple) applies the standard global BatchTopK constraint to the entire dictionary, including the core, matching the vanilla Matryoshka SAE setup for a more direct comparison. The sparse core DMSAE follows the same overall trends as dense core DMSAE. BatchTopK and Matryoshka SAE baselines from SAEBench are included for reference, and all SAEs were trained using a 65k sized dictionary.}
  \label{fig:saebench_core_sparsity_ablation}
\end{figure}

\clearpage
\section*{F. Pseudocode for DMSAE training and core selection}
This appendix provides pseudocode for (i) the attribution-guided core selection rule used to construct $C^{(t)}$ from the candidate pool $P^{(t)}$ (Algorithm~\ref{alg:select_core_attr}) and (ii) training a new DMSAE from a fixed distilled core $C^\star$ across sparsity targets (Algorithm~\ref{alg:dmsae_core_transfer}).

\begin{algorithm}[H]
\caption{\textsc{SelectCoreByAttribution}$(\theta,P,q,\tau,k,c)$}
\label{alg:select_core_attr}
\begin{algorithmic}[1]
\REQUIRE SAE parameters $\theta=\{W_{\mathrm{enc}},b_{\mathrm{enc}},W_{\mathrm{dec}},b_{\mathrm{dec}}\}$;
candidate pool $P$ (core+prefix-0); quantile $q$; coverage $\tau$;
\\ sparsity target $k$; core size $c$.
\ENSURE Ordered selected set $C\subseteq P$ for the next cycle.

\STATE Sample token positions $U$; obtain $(x_u,g_u)$ where $g_u=\partial L_{\mathrm{NT}}/\partial x_u$.
\FOR{each attribution batch $\{(x_u,g_u)\}$}
  \STATE Compute $f_u \leftarrow \mathrm{ReLU}(W_{\mathrm{enc}}x_u+b_{\mathrm{enc}})\in\mathbb{R}^{K}$.
  \STATE Set $\tilde f_u \leftarrow \big((f_u)_{0{:}c},\,\BatchTopK_k((f_u)_{c{:}K})\big)$.
  \FOR{each latent $j\in P$}
    \STATE $\bar w^{\mathrm{dec}}_j \leftarrow W_{\mathrm{dec}}[:,j]/\|W_{\mathrm{dec}}[:,j]\|_2$.
    \STATE For each $u$ in batch: compute $s_{u,j}\leftarrow g_u^\top \bar w^{\mathrm{dec}}_j$ and $\mathrm{GxA}_{u,j}\leftarrow |\tilde f_{u,j}s_{u,j}|$.
  \ENDFOR
\ENDFOR
\STATE For each $j\in P$, set $A_j \leftarrow \mathrm{Quantile}_q(\{\mathrm{GxA}_{u,j}\}_u)$.
\STATE Sort $P$ by descending $A_j$: $(j_1,\dots,j_{|P|})$; let $A_{\mathrm{tot}}=\sum_{j\in P}A_j$.
\STATE Return the smallest prefix $C=(j_1,\dots,j_r)$ s.t.\ $\sum_{\ell=1}^{r}A_{j_\ell}\ge \tau\,A_{\mathrm{tot}}$.
\end{algorithmic}
\end{algorithm}

\begin{algorithm}[H]
\caption{Training a DMSAE from a fixed distilled core $C^\star$ (core-transfer stage)}
\label{alg:dmsae_core_transfer}
\begin{algorithmic}[1]
\REQUIRE Distilled core encoder directions $\{w^\star_1,\dots,w^\star_c\}$ where $c=|C^\star|$;
total width $K$; \\ non-core prefix sizes $M=\{m_0<\cdots<m_L\}$; target sparsity $k$;
dataset $\mathcal{D}$; step size $\eta$.
\ENSURE Trained DMSAE parameters $\theta^\star\in\Theta^\star$.

\STATE Initialize parameters $\theta=\{W_{\mathrm{enc}},b_{\mathrm{enc}},W_{\mathrm{dec}},b_{\mathrm{dec}}\}$.
\STATE Set $W_{\mathrm{enc}}[0{:}c,:]\leftarrow [w^\star_1;\dots;w^\star_c]$ and freeze rows $0{:}c$.
\STATE Randomly initialize $W_{\mathrm{enc}}[c{:}K,:]$, all $b_{\mathrm{enc}}$, all $W_{\mathrm{dec}}$, and $b_{\mathrm{dec}}$.
\STATE Define feasible set $\Theta^\star \coloneqq \{\theta:\; W_{\mathrm{enc}}[0{:}c,:]=[w^\star_1;\dots;w^\star_c]\}$.

\STATE Set $k_{\text{non-core}} \leftarrow \operatorname{round}\!\left(k\cdot\frac{K-c}{K}\right)$.

\FOR{each mini-batch $X$ from $\mathcal{D}$}
  \STATE Compute $F \leftarrow f(X;\theta)\;=\;\mathrm{ReLU}(W_{\mathrm{enc}}X^\top + b_{\mathrm{enc}})^\top$.
  \STATE Set $\tilde F \leftarrow \big(F_{:,0{:}c},\,\BatchTopK_{k_{\text{non-core}}}(F_{:,c{:}K})\big)$.
  \STATE $\hat{\mathcal{J}}(X;\theta) \leftarrow \sum_{m\in M}\|X-\hat X_m(X;\theta,\tilde F)\|_F^2+\alpha\,\Laux(X;\theta,\tilde F)$.
  \STATE $\theta \leftarrow \Pi_{\Theta^\star}\!\left(\theta - \eta\,\nabla_\theta \hat{\mathcal{J}}(X;\theta)\right)$.
\ENDFOR
\STATE \textbf{Return:} $\theta^\star \leftarrow \theta$.
\end{algorithmic}
\end{algorithm}

\end{document}